\documentclass[final]{cvpr}
\usepackage{times}
\usepackage{epsfig}
\usepackage{graphicx}
\usepackage{amsmath}
\usepackage{amssymb}
\usepackage{multirow}
\usepackage{verbatim}
\usepackage{threeparttable}
\usepackage{caption}
\usepackage{wrapfig}
\usepackage{fancynum}
\usepackage{booktabs}
\usepackage{subcaption}
\usepackage[table, dvipsnames]{xcolor}

\usepackage[pagebackref=true,breaklinks=true,colorlinks,bookmarks=false]{hyperref}
\def\cvprPaperID{7060} 
\def\confYear{CVPR 2021}

\begin{document}

\title{NExT-QA: Next Phase of Question-Answering to Explaining Temporal Actions}

\author{Junbin Xiao, Xindi Shang, Angela Yao, Tat-Seng Chua \\
Department of Computer Science, National University of Singapore \\
{\tt\small{\{junbin,shangxin,ayao,chuats\}@comp.nus.edu.sg}}
}

\maketitle
\begin{abstract}
We introduce NExT-QA, a rigorously designed video question answering (VideoQA) benchmark to advance video understanding from describing to explaining the temporal actions. Based on the dataset, we set up multi-choice and open-ended QA tasks targeting causal action reasoning, temporal action reasoning, and common scene comprehension. Through extensive analysis of baselines and established VideoQA techniques, we find that top-performing methods excel at shallow scene descriptions but are weak in causal and temporal action reasoning. Furthermore, the models that are effective on multi-choice QA, when adapted to open-ended QA, still struggle in generalizing the answers. This raises doubt on the ability of these models to reason and highlights possibilities for improvement. With detailed results for different question types and heuristic observations for future works, we hope NExT-QA will guide the next generation of VQA research to go beyond superficial description towards a deeper understanding of videos.
(The dataset and related resources are available at \href{https://github.com/doc-doc/NExT-QA.git}{https://github.com/doc-doc/NExT-QA.git})
\end{abstract}

\section{Introduction}
\begin{figure}[t]
    \centering
    \includegraphics[width=0.48\textwidth]{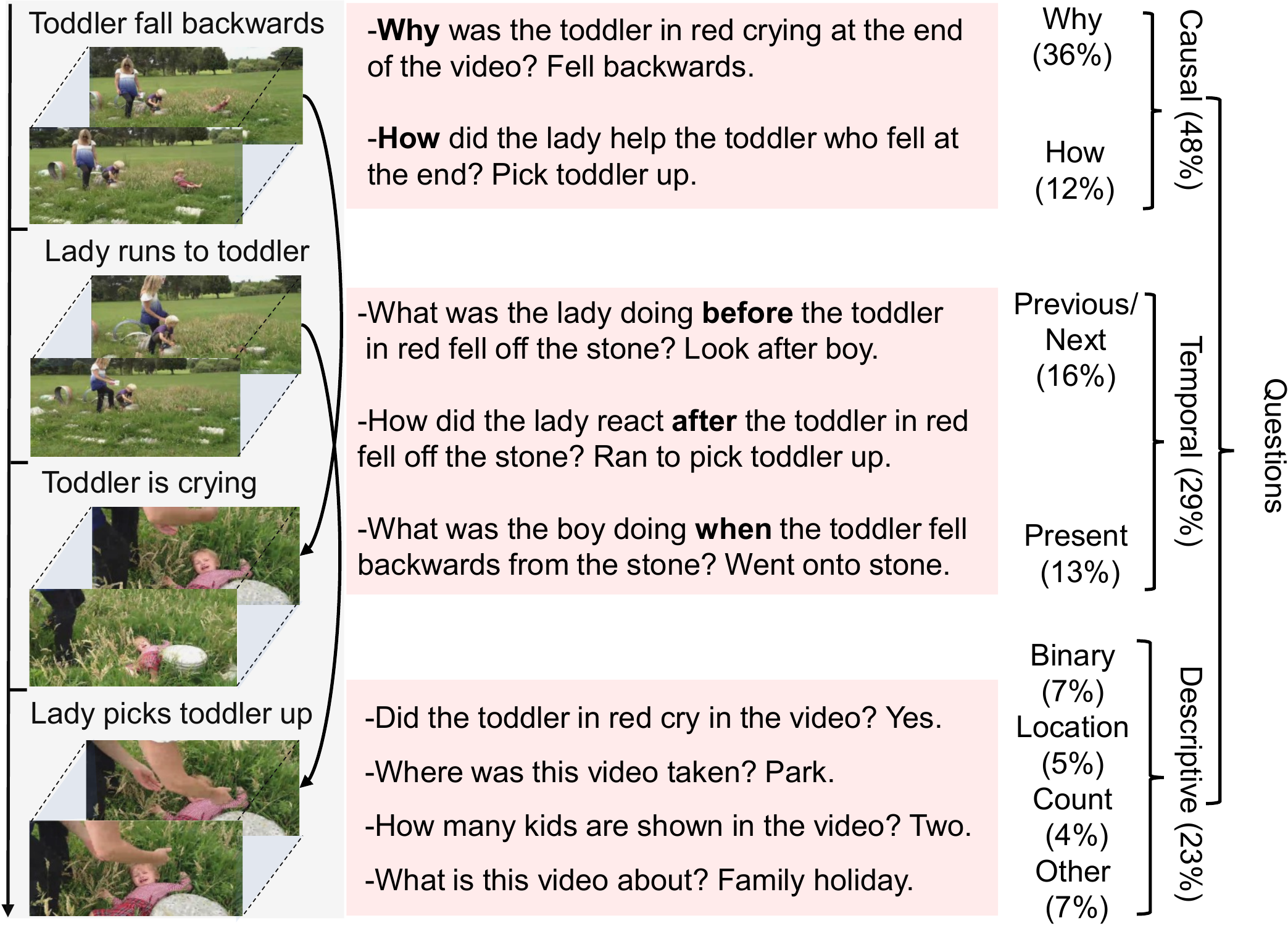}
    \caption{NExT-QA is a question answering benchmark
    targeting the explanation of video contents. It challenges QA models to reason about causal and temporal actions and understand the rich object interactions in daily activities.}
    \label{fig:intr}
\end{figure}

Actions in videos are often not independent but rather related with causal and temporal relationships \cite{buchsbaum2015inferring}. For example, in the video in Figure~\ref{fig:intr}, \emph{a toddler cries because he falls}, and \emph{a lady runs to the toddler in order
to pick him up}. Recognizing the objects ``\emph{toddler}'', ``\emph{lady}'' and describing the independent action contents like
``\emph{a toddler is crying}'' and ``\emph{a lady picks the toddler up}''
in a video are now attainable with advanced neural network models~\cite{he2016deep,lin2019tsm,zhou2018end}.
Yet being able to reason about their causal and temporal relations and answer natural language questions (\eg, ``\emph{Why is the toddler crying?}'', ``\emph{How did the lady react after the toddler fell?}''), which lies at the core of human intelligence \cite{shoham1988reasoning}, remains a great challenge for computational models and is also much less explored by existing video understanding tasks~\cite{krishna2017dense,SenerECCV2020,xiao2020visual,xu2016msr,yu2019activitynet}.

In this work, we study causal and temporal action reasoning in video question answering (VideoQA) and contribute NExT-QA, a benchmark to foster the \textbf{N}ext generation of VQA models to \textbf{Ex}plain \textbf{T}emporal actions. NExT-QA contains\fnum{5440}videos and about $52K$ manually annotated question-answer pairs
grouped into causal, temporal and descriptive questions. An overview of the typical questions and their distributions are found in Figure \ref{fig:intr}. To embody the reasoning challenges and provide effective diagnostics for video QA models, we set up two tasks at different difficulty levels. At the first level, multi-choice QA provides five candidate answers for each question and requires the models to pick out the correct one. At the second level, open-ended QA requires the models to generate the answers in short phrases with cues only from the videos and the questions (\ie, no candidate options).

Using NExT-QA, we evaluate several state-of-the-art (SOTA) video QA techniques \cite{fan2019heterogeneous,gao2018motion,jang2017tgif,jiang2020reasoning,le2020hierarchical,li2019beyond}. While the top-performing methods achieve compelling results on commonly descriptive questions, their performances on causal and temporal questions are far from satisfactory. Furthermore, when adapting the models that are effective on multi-choice QA to open-ended QA, we find that they struggle to automatically answer the questions.  This prompts a fundamental concern that these models do not truly understand the causal and temporal structure over the actions. As such, NExT-QA offers new challenges and ample opportunities to spark future research for a deeper understanding of video content.

To summarize our contributions: 1) we explore causal and temporal action reasoning in VideoQA to advance video understanding beyond shallow description towards deeper explanation; 2) we contribute NExT-QA, a rigorously curated VideoQA benchmark with manual annotations to foster research on causal and temporal action reasoning; and 3) we extensively analyze the baselines as well as the established video reasoning techniques on NExT-QA, providing detailed results for different question types and heuristic observations for future works.

\section{Related Work}
\textbf{Benchmarks}.
Early VideoQA benchmarks~\cite{maharaj2017dataset,xu2017video,xue2017unifying,ye2017video,zeng2017leveraging,zhu2017uncovering} rely on video descriptions \cite{li2016tgif,xu2016msr} (\eg, \emph{a man is skiing down a slope.}) to automatically generate question-answer pairs (\eg, \emph{Who is skiing down a slope? A man.}). They rarely require going beyond a recognition of the objects and actions to answer the questions. TGIF-QA \cite{jang2017tgif,jang2019VideoQA}, in particular, challenges spatio-temporal reasoning in animated GIFs.  However, GIFs are short videos (about 3s), and the actions are mostly trivial in describing the repetition or transition of a single object. Moreover, the questions are automatically populated from simple sentence templates. Consequently, SOTA methods~\cite{jang2019VideoQA,jiang2020reasoning,le2020hierarchical} perform well, leading to an inflated optimism of machine intelligence in video understanding. Recently, ActivityNet-QA~\cite{yu2019activitynet} was manually annotated to understand longer web videos. Yet, it has the same problems as in TGIF-QA, \ie, lacking object interactions and causal relationships.

Social-IQ \cite{zadeh2019social} is a newly proposed benchmark for social intelligence understanding. Although it is rich in causalities and interactions, it is small-scale and focuses on comprehending complex human social behaviours from multiple modalities (video, transcript, and audio).
Our dataset is larger and targets a richer set of causal and temporal actions in daily life, extending beyond human-social behaviours (\eg, \emph{The dog barks at the cat because the cat paws at the dog.}). Also, it requires videos as the only information source. MovieQA \cite{tapaswi2016movieqa} and TVQA \cite{lei2018tvqa} may also invoke causal and temporally related questions. Nonetheless, they are either biased to textual plot understanding or actor dialogue comprehension~\cite{dur31600}, severely diminishing their challenge for visual reasoning. More recently, CLEVRER~\cite{yi2019clevrer} specially studied temporal and causal relationships of physical motions in simulated environments. Our dataset is essentially different in that we explore causal and temporal actions for a deeper understanding of real-world videos.

Other works like Motivation \cite{vondrick2016predicting}, VCR \cite{Zellers2019CVPR} and V2C \cite{fang2020video2commonsense} may also take causality into consideration, either for visual description or QA. Nonetheless, they emphasize commonsense to imagine the predictions. Our work differs in that we focus on understanding the causal and temporal structure of the actions.  Specifically, we ensure that the answers to the questions are found in the video contents, \eg, for causal questions, we make sure that both the cause and effect actions are visible.
Such a setting is impossible in static images \cite{vondrick2016predicting,Zellers2019CVPR} that requires models to speculate or make commonsense reasoning, which goes in an orthogonal direction to our aim. Finally, we note that QA on causal and temporal events have long been studied in text comprehension \cite{girju2003automatic,ning2018joint}. However, these works focus on detecting lexico-syntactic patterns that express causation on news events rather than reasoning over specific videos' causal/temporal actions.

\textbf{Techniques}.
Language-guided visual reasoning like VQA has progressed significantly driven by the tremendous advancements in object/action recognition \cite{carreira2017quo,hara2018can,he2016deep,simonyan2014two,tran2015learning} and natural language understanding \cite{cho2014learning,devlin2018bert,hochreiter1997long,pennington2014glove,vaswani2017attention}. Most of the improvements have been made in image QA \cite{anderson2018bottom,antol2015vqa,lu2016hierarchical} though video QA has received increasing attention recently. Established works \cite{jang2017tgif,tapaswi2016movieqa,zeng2017leveraging,xu2017video} apply 2D convolutional neural networks (CNNs) (\eg, ResNet \cite{he2016deep}) to learn frame-level appearance feature, and 3D CNNs (\eg, C3D \cite{tran2015learning}, I3D \cite{carreira2017quo,hara2018can}) or optical flow to capture clip-level (or segment-level) motion information. The final video-level representation can be obtained by simple pooling or more sophisticated aggregation models, such as temporal relation networks (\eg, TCN \cite{lea2017temporal}, TRN \cite{zhou2018temporal} and CRN \cite{le2020hierarchical}), sequential models (\eg, RNNs with LSTM \cite{hochreiter1997long}, GRU \cite{cho2014learning} and their variants) and attention \cite{li2019beyond,jin2019multi}. During aggregation, the textual clues from the question side (usually modeled by RNNs) are integrated for language-guided video reasoning and are achieved by additional reasoning modules, such as spatial and temporal attention \cite{jang2019VideoQA, jang2017tgif,jin2019multi,zhao2017video}, co-attention \cite{jiang2020reasoning,li2019beyond,lu2016hierarchical}, multi-cycle memory \cite{fan2019heterogeneous,gao2018motion}, graph neural networks \cite{huang2020location,jiang2020reasoning} and conditional relation networks \cite{le2020hierarchical}. In this work, we will comprehensively analyze the relevant techniques on NExT-QA, providing effective baselines and heuristic observations.

\section{NExT-QA Dataset}
\subsection{Criteria and Task Definition}
\textbf{Causal Questions} are designed to explain actions, either uncovering the intentions of the previously occurring actions or stating causes for subsequent actions. In this work, \emph{`A explains B'} of two actions A and B in a given video means that A is a visible cause responsible for B's occurrence. Thus, questions in the causal group ask either why the objects act in a certain manner or how (what did they do) to bring about an observed effect. Accordingly, both causes and effects should be visible in the videos. Examples can be found in Figure \ref{fig:intr} (top).

\textbf{Temporal Questions} assess the model's capability of reasoning about temporal relationships between actions. Temporal actions, while related to causality, are determined only by order of occurrence. Hence, questions of this type ask about the previous (\emph{what ... do before ...}), present (\emph{what ... doing when/while/as ...}) or next actions (\emph{what/how ... do/react after ...}). Unlike previous works \cite{jang2017tgif, yu2019activitynet} which focus on reasoning temporal actions of a single object in a question, we emphasize more on object interactions. Examples can be found in Figure \ref{fig:intr} (middle).

\textbf{Descriptive Questions} focus on scene description of the videos (\eg, the places, objects / attributes, and main actions / events). These questions complement causal and temporal questions to make up a holistic video comprehension and also allow for comparison between different types of questions. Specifically, the questions cover binary choices (yes/no, or the answers are indicated in the questions, \eg, ``\emph{... tired or energetic?}"), location (where), counting (how many) and other free-form questions. The only requirement for free-form questions is that the answers can be visibly inferred from videos and are not subjective. Examples can be found in Figure \ref{fig:intr} (bottom).

\textbf{Multi-choice vs. Open-ended QA}. We define two tasks based on the above question types. In multi-choice QA, models are presented with five options (one correct answer plus four distractor answers) from which they are required to select the correct one. Providing candidate answers brings convenience in prediction evaluation. However, it diminishes the reasoning challenge, as models are prone to learn the difference between the correct and incorrect answers purely; this is especially the case when the wrong answers are not generated to be challenging enough. Also, it dispenses with the need for answer generation, which in our view should be an interesting and open field of research in QA. Therefore, we also study open-ended QA where no candidate answers are provided, and the models must interpret the question and video contents and generate the textual answers automatically. Previous works~\cite{antol2015vqa,jang2017tgif,yu2019activitynet} formulate open-ended QA as a classification problem to classify the video-question pairs into a fixed answer set.  We set it as a generation problem since the answers are mostly simple phrases in NExT-QA. Generation-based open-ended QA is of higher practical value and also receives widespread attention recently \cite{xue2017unifying,zhao2017video,zhao2018open}.

\subsection{Dataset Construction}
\textbf{Video Source}. We aimed for natural videos featuring object interaction in daily life, without restriction to certain actors and activities.
With these goals in mind, we found the video relation dataset VidOR \cite{shang2019annotating}~\footnote{Videos are drawn from YFCC-100M \cite{thomee2016yfcc100m} and are crawled from Flickr.} suits our requirements well. We selected from VidOR \fnum{6000} videos that are longer and richer in objects and interactions. Although we do not restrict the content, the videos are mainly about family time, kids playing, social gatherings, outdoor activities, pets and musical performances.  We randomly split the videos into train/val/test sets with a ratio of 7:1:2.

\textbf{Annotation} of the NExT-QA dataset was done in 3 stages\footnote{Annotating all the questions in one stage was problematic for quality control and compensation. We annotated first the causal questions and then temporal; descriptive questions were the easiest and done last. Payment was commensurate with the number and difficulty of the questions.} over one year by 100 undergraduate students. The annotators were supervised at each stage with the following principles to ensure high-quality annotations. 1) All the annotators are rigorously trained before doing the actual annotation. 2) Question and answer annotation are done by separate annotators. Answer annotators are expected to check the questions' quality first, answer the good questions and fix (or delete) the bad ones. In this way, we can simulate the evaluation process and ensure that the questions are answerable and not subjective. 3) Suggested maximal lengths for questions and answers are 22 and 6 words, respectively. We especially encouraged succinct answers to avoid sentence paraphrasing and to facilitate answer evaluation. 4) The question types are set in a drop-down menu and must be selected by the questioners to ensure the distribution of the questions satisfying each video's requirements. 5) Questioners can report videos that are hard to pose effective questions. The confirmed boring videos are removed from the database.

\textbf{Post-Processing}. We removed some \emph{yes}-answered questions in the validation and test sets to ensure a balanced number of answers for \emph{yes} and \emph{no}. Additionally, we deleted a limited number of counting questions whose answer values are larger than twenty. What remained are\fnum{5440}valid videos and\fnum{52044}question-answer pairs; detailed statistics are presented in Sec. \ref{sec:data-stat}.

\textbf{Multi-choice Generation}.
To be meaningful, the distractors in multi-choice QA should be unique to each other, semantically coherent in answering the questions, and different in meaning with respect to the correct answer. To this end, we first grouped the questions according to the annotated question types (binary questions are excluded). Then, for each question, we retrieved the top 50 questions similar to the queried question in the same group according to their cosine similarities based on off-the-shelf features of Sentence-BERT \cite{reimers-2019-sentence-bert}. The answers to these 50 questions are returned as distractor candidates and then filtered for redundancy and similarity to the correct answer. Two answers are redundant or similar if 1) their lemmatized variants are the same, in which stop words are not considered, or 2) the cosine similarity of their feature vectors is large than 0.9. To ensure hard negatives, we also discard the candidate whose similarity with the correct answer is lower than 0.2. Afterwards, we sampled four qualified candidates as distracting answers for each question and randomly (but evenly) insert the correct answers to form 5 options. Finally, we manually checked all the question-answer tuples and amended some options to ensure the effectiveness of the generated multiple choices. We show some examples in Figure \ref{fig:multi-exp-s}; more are found in the Appendix.

\begin{figure}[t]
    \centering
    \includegraphics[width=0.48\textwidth]{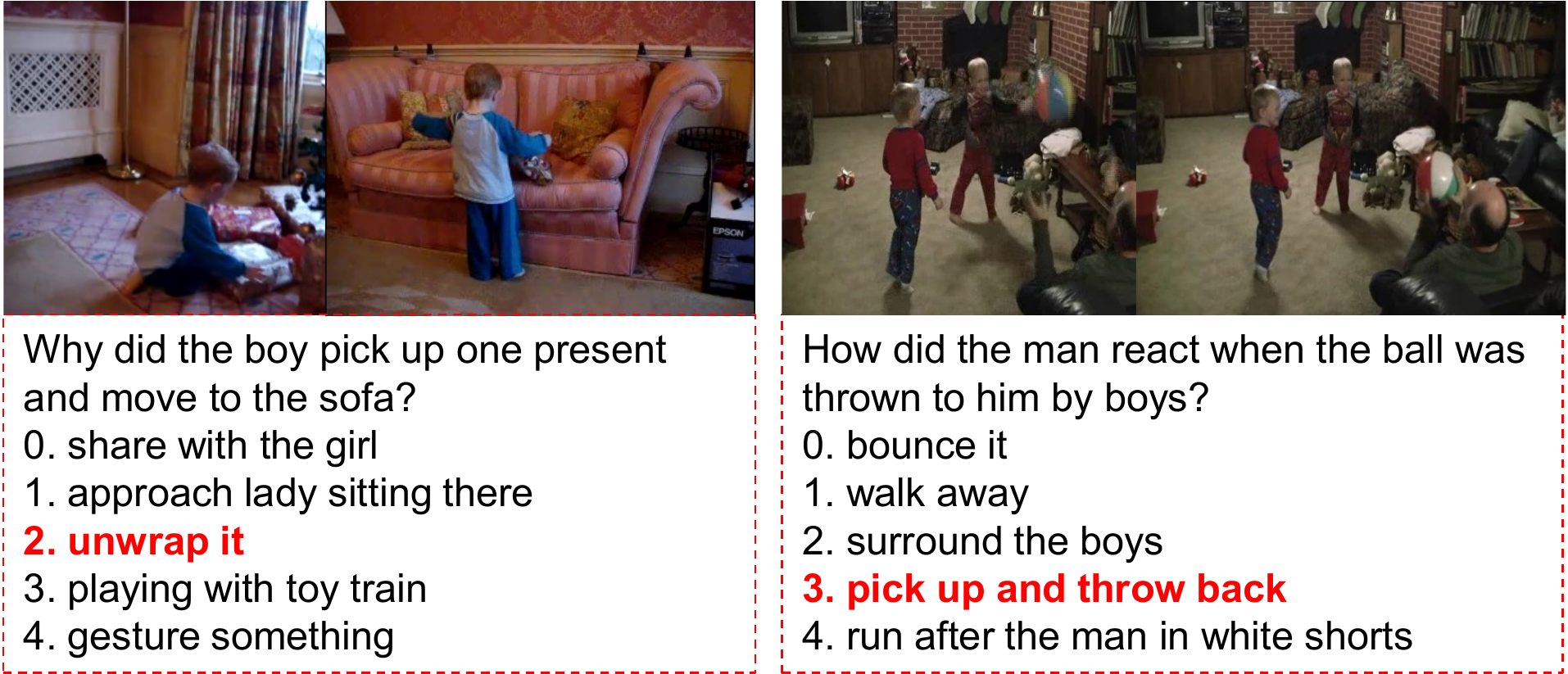}
    \caption{Examples of multi-choice QA.}
    \label{fig:multi-exp-s}
\end{figure}

\setlength{\belowcaptionskip}{-0.5cm}
\setlength{\tabcolsep}{2pt}
\begin{table}[t]
\small
\centering
\begin{threeparttable}
    \scalebox{0.8}{
    \begin{tabular}{ccccccccc}
        \toprule
         \multicolumn{4}{c}{Videos}&\multirow{2}{*}{Tasks} & \multicolumn{4}{c}{Questions} \cr
         \cmidrule(lr){1-4}  \cmidrule(lr){6-9}
        Train & Val & Test &Total & & Train & Val & Test & Total\\
        \midrule
        \midrule
          \multirow{2}{*}{\fnum{3870}} & \multirow{2}{*}{570} & \multirow{2}{*}{\fnum{1000}} & \multirow{2}{*}{\fnum{5440}}& Multi-Choice QA & \fnum{34132} & \fnum{4996} & \fnum{8564} & \fnum{47692}\cr
           &  &  &  &Open-Ended QA & \fnum{37523} & \fnum{5343} & \fnum{9178} & \fnum{52044}\cr
        \bottomrule
    \end{tabular}
    }
    \caption{Statistics of the NExT-QA datasets.}
    \label{tab:datasets}
\end{threeparttable}
\end{table}

\subsection{Data Statistics}
\label{sec:data-stat}
\begin{figure}[t]
    \centering
    \includegraphics[width=0.485\textwidth]{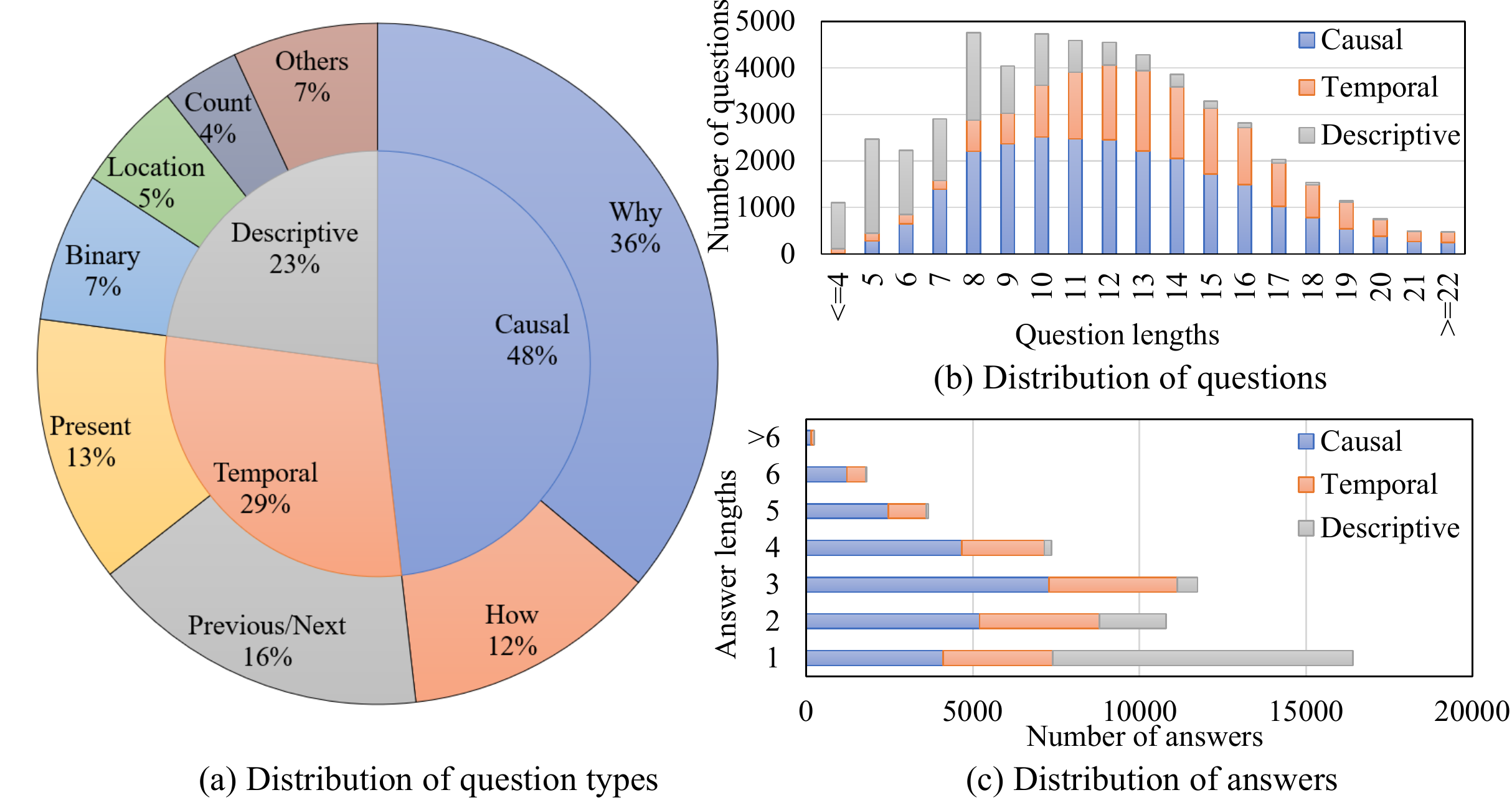}
    \caption{Data statistics. (a) Distribution of the question types. (b) The average question length is 11.6, and the specific lengths for causal, temporal and descriptive questions are 12.1, 13.4 and 8.0 respectively. (c) The average answer length is 2.6. Specific lengths for causal, temporal and descriptive answers are 3, 2.8 and 1.4 respectively.}
    \label{fig:data-statistic}
\end{figure}

NExT-QA contains\fnum{5440}videos, including\fnum{3870}for training, 570 for validation and\fnum{1000}for testing. Detailed statistics are given in Table \ref{tab:datasets}. The distribution of the questions and answers are shown in Figure \ref{fig:data-statistic}. From Figure \ref{fig:data-statistic}(a) we can see that the number of causal questions accounts for approximately half (48\%) of the whole dataset; questions starting with 'why' are the majority, constituting 36\%.
Temporal questions of understanding the present or inferring the past or future compose 29\% of the whole dataset. Apart from causal and temporal questions, there is 23\% of descriptive questions which focus on describing the locations, objects/attributes and main events in the videos.
\setlength{\belowcaptionskip}{-0.5cm}

\setlength{\belowcaptionskip}{-0.5cm}
\begin{table*}[t]
\small
\centering
\begin{threeparttable}
    \scalebox{1.0}{
    \begin{tabular}{lccccccc}
        \toprule
             Dataset & Video Source & Goal & Annotation & \#Videos & \#QA Pairs & Video Length (s) & QA Task \cr
        \midrule
        \midrule
         MSVD-QA~\cite{xu2017video}& MSVD & descriptive QA & Auto  & \fnum{1970} & \fnum{50505} & 10 & OC\cr
         MSRVTT-QA~\cite{xu2017video}& MSRVTT & descriptive QA & Auto  & \fnum{10000} & \fnum{243690} & 15& OC\cr
         TGIF-QA~\cite{jang2019VideoQA,jang2017tgif}& TGIF & spatio-temporal reasoning & Auto  & \fnum{71741} & \fnum{165165} & 3 & MC\&OC\cr
         TVQA~\cite{lei2018tvqa}&TV Show & subtitle\&concept comprehension & Man  & \fnum{21793} & \fnum{152545} & 76 & MC\cr
         ActivityNet-QA~\cite{yu2019activitynet}&ActivityNet & descriptive QA & Man  & \fnum{5800} & \fnum{58000} & 180 & OC\cr
         Social-IQ~\cite{zadeh2019social}&YouTube & social intelligence understanding & Man  & \fnum{1250} & \fnum{7500} & 60 & MC\cr
         \midrule
         NExT-QA (ours) & YFCC-100M & causal \& temporal action interactions & Man  & \fnum{5440} & \fnum{52044} & 44 & MC\&OG \cr
        \bottomrule
    \end{tabular}
    }
    \caption{Dataset comparisons. OC and OG denote \textbf{O}pen-ended question-answering as problem of \textbf{C}lassification and \textbf{G}eneration respectively. MC stands for multi-choice QA.
    }
\label{tab:dataset_comp}
\end{threeparttable}
\end{table*}

The distribution of question word length is shown in Figure \ref{fig:data-statistic}(b). Questions are on average 11.6 words, which is much longer than existing VideoQA datasets (\eg, 8.7 in Activity-QA \cite{yu2019activitynet}). We find a clear difference in the three question types' distributions, \ie, descriptive questions are the shortest while questions for causal and temporal actions are relatively longer. This is reasonable as most of the descriptive questions have a simple syntactic structure, while the questions in the causal and temporal groups are mostly compounded. Accordingly, answers (Figure \ref{fig:data-statistic}(c)) to the descriptive questions are shorter since they are related to video recognition. In contrast, answers to causal and temporal questions are relatively longer. Nevertheless, the vast majority of the questions can be answered in 6 words.

\subsection{Dataset Comparison}
NExT-QA has several attractive properties compared with other datasets (see Table \ref{tab:dataset_comp}; a more detailed analysis is given in Appendix part 1).  First, NExT-QA is unique in that it goes beyond descriptive QA to benchmark causal and temporal action reasoning in realistic videos and is also rich in object interactions. Second, it is among the largest VideoQA datasets that are manually annotated to support both multi-choice and open-ended QA, allowing comprehensive comparisons of different VQA techniques. Finally, the videos in NExT-QA are rich and diverse in terms of objects, actions and events, and all reflect real daily life, which differs from the popular TVQA \cite{lei2018tvqa} dataset that biased towards comprehending dialogues between main characters in the TV shows.

\section{Experiments}

\textbf{Evaluation}. For multi-choice QA, we report the accuracy or percentage of correctly answered questions. For open-ended QA, we first remove the stop words and lemmatize other words in the answers. Then, we determine the Wu-Palmer similarity (WUPS) score \footnote{WUPS computes the Wu-Palmer similarity \cite{wu1994verbs} of the words based on their depths in WordNet \cite{miller1995wordnet}. It can be regarded as a soft version of accuracy that factors in synonyms and other semantics~\cite{xue2017unifying,zhao2018open}}~\cite{malinowski2014multi} to evaluate the quality of the generated answers. For binary and counting questions in the descriptive group, we use accuracy instead. Since accuracy is easily integrated into WUPS (as a hard version), we do not report them separately for brevity.

\textbf{Configuration}. We uniformly sample for each video 16 clips (segments), and each has 16 consecutive frames. The per-frame appearance feature is extracted from ResNet-101 \cite{he2016deep} pretrained on ImageNet \cite{deng2009imagenet}, from either Convolutional (Conv) layers or fully connected (FC) layers depending on the specific models. The clip-level motion information is captured by inflated 3D ResNeXt-101 \cite{hara2018can,xie2017aggregated} pre-trained on Kinetics \cite{kay2017kinetics}. On the language side, we study both GloVe \cite{pennington2014glove} for word representations as in the original paper and the recent BERT \cite{devlin2018bert} for sentence embedding. Unless otherwise indicated, for multi-choice QA, the candidate answers are concatenated to the corresponding questions, and the models are optimized by maximizing the margins between the correct and incorrect QA pairs using Hinge loss. For open-ended QA, the video-question communicated features will be fed to the answer decoders to generate the answers word by word. The models are optimized by minimizing the softmax cross-entropy loss. All the experiments follow the data split in Table \ref{tab:datasets}. We train the models on the respective training sets, during which the optimal model settings are explored on the validation sets.

\subsection{Multi-choice QA}
We first discuss the baselines designed to diagnose any potential biases in NExT-QA and then analyze the established video reasoning techniques.
\vspace{-0.3cm}
\subsubsection{Baselines}
\textbf{Random}. This baseline randomly chooses one option as the correct answer and keeps it the same for all the questions. Table \ref{tab:bs_results} shows the results of always selecting the first option as a representative. The random accuracy across different question types is about 20\%, as the correct answers are evenly distributed among the five options.

\textbf{Longest}, \textbf{Shortest} and \textbf{Popular}. As the names suggest, the \emph{longest} / \emph{shortest} baselines always select the longest / shortest answer as the correct one. We can see that both methods improve little over the random baseline. When we regulate the strategy a bit by selecting the most popular answers (\ie, the most frequent answer for each question type) if it is among the five options otherwise choosing the shortest one,
as shown in the \emph{Pop.+Short} baseline, there are clear improvements for questions in the descriptive group. Yet, the results are only slightly better for causal questions and even worse for temporal questions. This is understandable as descriptive questions are easier to have frequent answers.  All these baselines verified that an educated guess is hard to achieve good results on NExT-QA.

\setlength{\belowcaptionskip}{-0.675cm}
\setlength{\tabcolsep}{7pt}
\begin{table}[t]
\begin{center}
\scalebox{0.9}{
\begin{tabular}{lccccc}
\toprule
Methods & Text Rep. & $Acc_C$ & $Acc_T$ & $Acc_D$ & $Acc$ \cr
\midrule
\midrule
Random & -& 20.52 & 20.10 & 19.69 & 20.08\cr
Longest & - & 21.71 & 21.46 & 17.89 & 21.04\cr
Shortest & - & 22.09 & 19.67 & 22.78 & 21.42\cr
Pop.+Short & - & 22.25 & 20.41 & 32.43 & 23.24\cr
SimAA & Se-BERT &18.11 & 19.23 & 18.15 & 18.47\cr
SimQA & Se-BERT &27.12 & 26.67 & 26.64 & 26.90 \cr
BlindQA & GloVe & 26.89 & 30.83 & 42.60 & 30.60\cr
BlindQA & BERT & 23.78 & 24.26 & 35.26 & 25.72\cr
BlindQA & BERT-FT & 42.62 & 45.53 & 43.89 & 43.76\cr
\rowcolor{lightgray}
Human &- & 87.61 & 88.56 & 90.40 & 88.38\cr
\bottomrule
\end{tabular}
}
\caption{Baseline and human results on validation set. $Acc_C$, $Acc_T$ and $Acc_D$ denote accuracy for causal, temporal and descriptive questions respectively.}
\label{tab:bs_results}
\end{center}
\end{table}

\setlength{\belowcaptionskip}{-0.675cm}
\setlength{\tabcolsep}{9pt}
\begin{table*}[t]
\begin{center}
\scalebox{0.8}{
\begin{tabular}{lcccccccccccc}
\toprule
\multirow{2}{*}{Methods} & \multirow{2}{*}{Text Rep.} & \multicolumn{3}{c}{$Acc_C$} & \multicolumn{3}{c}{$Acc_T$} & \multicolumn{4}{c}{$Acc_D$} & \multirow{2}{*}{$Acc$}\cr
\cmidrule(lr){3-5} \cmidrule(lr){6-8} \cmidrule(lr){9-12}
& & Why & How & All & Prev\&Next & Present & All & Count & Location & Other & All\cr
\midrule
\midrule
EVQA \cite{antol2015vqa} & GloVe & 28.38 & 29.58 & 28.69 & 29.82 & 33.33 & 31.27 & 43.50 & 43.39 & 38.36 & 41.44 & 31.51\cr
PSAC \cite{li2019beyond} & GloVe & 35.81 & 29.58 & 34.18 & 28.56 & 35.75 & 31.51 & 39.55 & 67.90 & 35.41 & 48.65 & 35.57\cr
PSAC+ \cite{li2019beyond} & GloVe & 35.03 & 29.87 & 33.68 & 30.77 & 35.44 & 32.69 & 38.42 & 71.53 & 38.03 & 50.84 & 36.03\cr
CoMem \cite{gao2018motion} & GloVe & 36.12 & 32.21 & 35.10 & 34.04 & 41.93 & 37.28 & 39.55 & 67.12 & 40.66 & 50.45 & 38.19\cr
STVQA \cite{jang2017tgif} & GloVe& 37.58 & 32.50 & 36.25 & 33.09 & 40.87 & 36.29 & \underline{45.76} & 71.53 & 44.92 & 55.21 & 39.21\cr
HGA \cite{jiang2020reasoning} & GloVe & 36.38 & 33.82 & 35.71 & 35.83 & 42.08 & 38.40 & {\bf46.33} & 70.51 & 46.56 & 55.60 & 39.67\cr
HME \cite{fan2019heterogeneous} & GloVe & 39.14 & 34.70 & 37.97 & 34.35 & 40.57 & 36.91 & 41.81 & 71.86 & 38.36 & 51.87 & 39.79\cr
HCRN \cite{le2020hierarchical} & GloVe & 39.86 & 36.90 & 39.09 & 37.30 & 43.89 & 40.01 & 42.37 & 62.03 & 40.66 & 49.16 & 40.95\cr
\midrule
EVQA \cite{antol2015vqa} & BERT-FT & 42.31 & 42.90 & 42.46 & 46.68 & 45.85 & 46.34 & 44.07 & 46.44 & 46.23 & 45.82 & 44.24\cr
STVQA \cite{jang2017tgif} & BERT-FT& 45.37 & 43.05 & 44.76 & 47.52 & \underline{51.73} & \underline{49.26} & 43.50 &65.42 & \underline{53.77} & 55.86 & 47.94\cr
CoMem \cite{gao2018motion} & BERT-FT & 46.15 & 42.61 & 45.22 & \underline{48.16} & 50.38 & 49.07 & 41.81 & 67.12 & 51.80 & 55.34 & 48.04\cr
HCRN* \cite{le2020hierarchical} & BERT-FT & {\bf 46.99} & 42.90 & 45.91 & \underline{48.16} & 50.83 & \underline{49.26} & 40.68 & 65.42 & 49.84 & 53.67 & 48.20\cr
HME \cite{fan2019heterogeneous} & BERT-FT & \underline{46.52} & {\bf45.24} & \underline{46.18} & 47.52 & 49.17 & 48.20 & 45.20 & {\bf73.56} & 51.15 & \underline{58.30} & \underline{48.72}\cr
HGA \cite{jiang2020reasoning} & BERT-FT & {\bf46.99} & \underline{44.22} & {\bf46.26} & {\bf49.53} & {\bf52.49} & {\bf50.74} & 44.07 & \underline{72.54} & {\bf55.41} & {\bf59.33} & {\bf49.74}\cr

\bottomrule
\end{tabular}
}
\caption{Results of multi-choice QA on validation set. +: add motion feature. *: concatenate the question and answer to adapt to BERT representation. (The \textbf{best} and \underline{second best} results are bolded and underlined respectively.)}
\label{tab:mulresults}
\end{center}
\end{table*}

\textbf{SimAA} and \textbf{SimQA}. We specifically analyze the retrieval-based methods since the negative answers are mainly generated by searching the nearest neighbours of questions on the dataset. Concretely, the SimAA baseline is designed to check whether or not the correct answers are semantically far away from the distractor answers. To this end, we extract Sentence-BERT \cite{reimers-2019-sentence-bert} representation (Se-BERT) for the answers and find the furthest from the other four options as the correct answer for each question.

As shown in Table \ref{tab:bs_results}, this baseline performs the worst among all methods, revealing that the answers are challenging to disambiguate without seeing the questions and videos. Similarly, we design the SimQA baseline to retrieve the answers closest to the corresponding questions in the feature space. This baseline performs relatively better than the previously introduced baselines on causal and temporal questions, but its performance is still worse than the \emph{Popular+Shortest} baseline on the descriptive questions. The results are reasonable as there is less semantic overlap between the descriptive group's questions and answers. Again, these results suggest that the questions cannot be answered well simply based on semantic similarity between the questions and answers.

\textbf{BlindQA}. We study a blind version of deep models by considering the question-answers only and ignoring the video parts. To this end, we model the QAs with LSTM, during which the words are initialized with either GloVe \cite{pennington2014glove} or BERT \cite{devlin2018bert} representations. As a popular fashion, we extract token representations from the penultimate layer of the BERT-base model. As shown in Table \ref{tab:bs_results}, the BlindQA models steadily improve the results over all question types. Intriguingly, the model that utilizes GloVe performs better than that using BERT. We believe this is because the off-the-shelf BERT representations are seriously biased to the corpus on which it was trained and thus generalizes poorly to the scenario where the text is mostly visual-content related. Therefore, we further fine-tune BERT for multi-choice QA by maximizing the correct QA pairs' probability in each multi-choice QA. From Table \ref{tab:bs_results}, we can see that BERT-FT remarkably boosts the results over the off-the-shelf BERT representation and also GloVe. Nonetheless, the results are still much worse than human performance and thus indicate the necessity of understanding videos.

\vspace{-0.3cm}
\subsubsection{Established VideoQA Models}
We analyze and benchmark several established VideoQA methods in Table \ref{tab:mulresults} and Table \ref{tab:sota_results}, covering diverse network architectures and visual reasoning techniques.

EVQA \cite{antol2015vqa} extends the BlindQA baseline by adding up the visual stream, which is modelled by another LSTM. 
The visual and textual features are then element-wise added to predict the answers. Without any reasoning modules in the model, it trivially outperforms the BlindQA baseline.
STVQA \cite{jang2017tgif,jang2019VideoQA} advances EVQA by applying two dual-layer LSTMs for video and question modelling, with additional spatio-temporal attention modules for visual reasoning. We can see that it steadily boosts the EVQA baseline's performance across all 3 types of questions. The same is observed for CoMem \cite{gao2018motion} and HME \cite{fan2019heterogeneous}. Both share similar video and question encoders as in STVQA but use memory modules for visual appearance, motion and language reasoning in a multi-cycle fashion \footnote{We use the implementation provided by \cite{fan2019egovqa} as there is no official code available for CoMem. The video encoder is a two-layer GRU \cite{cho2014learning} instead of TCN \cite{lea2017temporal} used in the original paper.}.

Unlike the above methods that apply RNNs to contextualize video representations, PSAC \cite{li2019beyond} utilizes self-attention (the building block of transformer architectures \cite{vaswani2017attention}) on top of CNN feature and achieves great success on TGIF-QA \cite{jang2017tgif} with merely appearance feature. As the transformer essentially stacks fully-connected layers with short-cut connections, it trains fast but is data-hungry; on NExT-QA, it suffers from over-fitting problem and performs the worst among other methods. We speculate that the dataset is likely not large enough to learn transform-style visual models directly. Nevertheless, it would be a good testbed for pre-trained architectures~\cite{sun2019videobert,Zhu_2020_CVPR}.

HCRN \cite{le2020hierarchical} is a hierarchical model with conditional relation networks (CRN) as building blocks. It operates on video frame/segment sets of variable lengths conditioned on either motion or textual clues in a stage-wise fashion to reason on the video at multiple granularities. As shown in Table \ref{tab:mulresults}, it shows strong performance for causal and temporal action reasoning when GloVe representations are considered. However, when it is adapted to BERT representation, the results are not consistently good. Such difference could be that the size of the model is one order of magnitude larger than the others and is thus prone to be over-fitting as the size of BERT representation is approximately 2.5 times larger than that of GloVe (768 \vs 300).

HGA \cite{jiang2020reasoning} introduces a heterogeneous graph reasoning module and a co-attention unit to capture the local and global correlations between video clips, linguistic concepts and their cross-modal correspondence. The method is better suited for causal and temporal action reasoning and shows superior performance with BERT representations, achieving the SOTA results on NExT-QA. Yet, the gap between human performance remains large (\eg, 46.26\% \vs 87.61\% on causal questions, 50.74\% \vs 88.56\% on temporal questions, 59.33\% \vs 90.40\% on descriptive questions), and thus offers ample opportunity for improvement.

\setlength{\belowcaptionskip}{-0.675cm}
\setlength{\tabcolsep}{20pt}
\begin{table}[t]
\begin{center}
\scalebox{0.7}{
\begin{tabular}{lccccc}
\toprule
Methods & $Acc_C$ & $Acc_T$ & $Acc_D$ & $Acc$ \cr
\midrule
\midrule
EVQA \cite{antol2015vqa} &43.27 & 46.93 & 45.62 & 44.92\cr
STVQA \cite{jang2019VideoQA} & 45.51 & 47.57 & 54.59 & 47.64\cr
CoMem \cite{gao2018motion} & 45.85 & \textbf{50.02} & 54.38 & 48.54\cr
HCRN \cite{le2020hierarchical} & \underline{47.07} & \underline{49.27} & 54.02 & 48.89\cr
HME \cite{fan2019heterogeneous} & 46.76 & 48.89 & \underline{57.37} & \underline{49.16}\cr
HGA \cite{jiang2020reasoning} & \textbf{48.13} & 49.08 & \textbf{57.79} & \textbf{50.01}\cr
\bottomrule
\end{tabular}
}
\caption{Results of multi-choice QA on test set. All are based on fine-tuned BERT representation.}
\label{tab:sota_results}
\end{center}
\end{table}

\setlength{\belowcaptionskip}{-0.2cm}
\begin{figure}[t]
    \centering
    \includegraphics[width=0.48\textwidth]{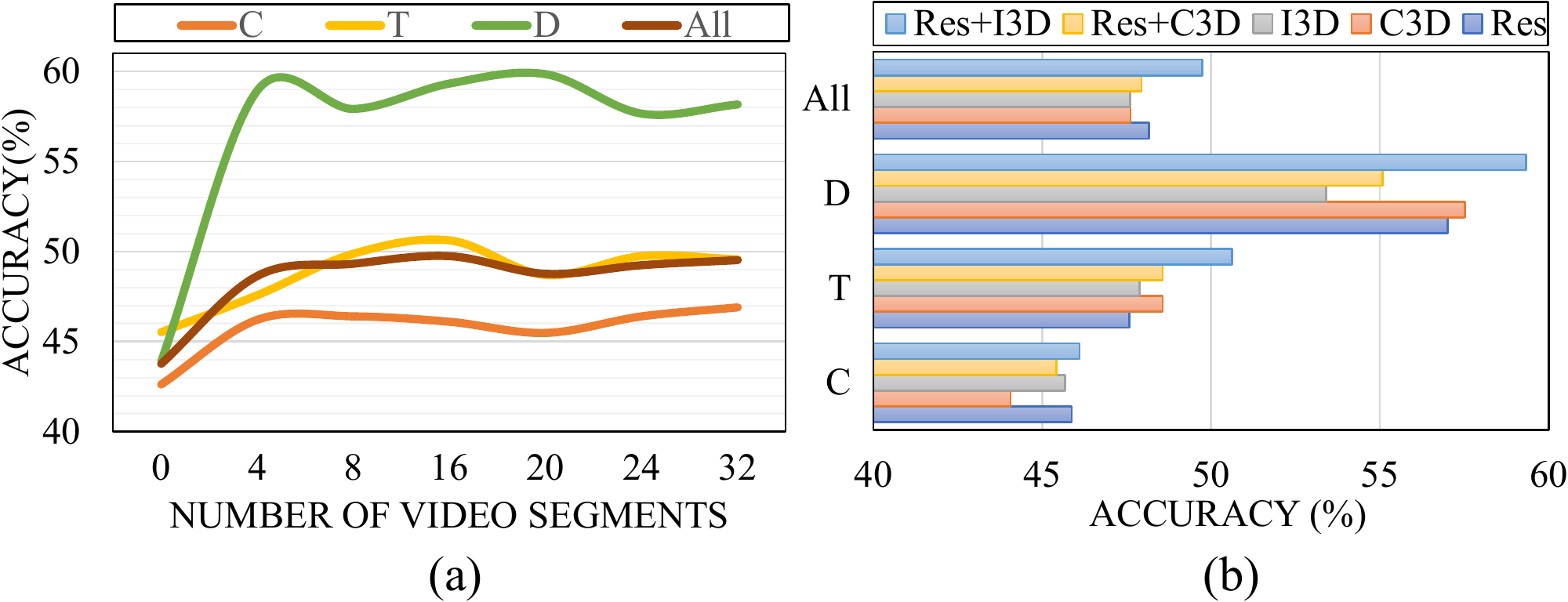}
    \caption{(a) Results with different number of clips. (b) Results with different video representations. C, T and D stand for causal, temporal and descriptive questions respectively.}
    \label{fig:res-mul-clip}
\end{figure}
\vspace{-0.3cm}
\subsubsection{Video Sampling Rates and Representations.}
\label{sec:sap-rep}

We based on HGA with BERT-FT as language representation to analyze the influence of video sampling rates and feature representations.
First, we vary the number of sampled video clips (segments) from 0 to 32, where 0 stands for the respective BlindQA baseline.
As shown in Figure \ref{fig:res-mul-clip}(a), we can see clear improvements for all types of questions with attendance of videos. Specifically, the improvement for descriptive questions is significant with more than 15\%. Besides, we also observe that 16 segments are enough to obtain good overall accuracy, whereas it needs relatively more segments to achieve better results on causal questions.

In Figure \ref{fig:res-mul-clip}(b), we investigate different features of video frames and segments. From the results, we can conclude that, for all types of questions, the best performance is from using ResNet as an appearance feature along with I3D ResNeXt as a motion feature (Res+I3D). When I3D is replaced with C3D (Res+C3D), results drop for all questions even though we do not observe absolute weakness between C3D and I3D in this experiment. We speculate that improvements can be mainly attributed to 1) I3D performing better on causal questions which account for the majority of NExT-QA; and 2) ResNeXt which was derived from ResNet and thus matches better with ResNet in feature space than C3D. A similar observation was made in \cite{jang2019VideoQA}.

\subsection{Open-ended QA}
We transfer several top-performing methods in multi-choice QA to open-ended QA. To this end, we first build a vocabulary set of \fnum{3392} words by selecting those appearing more than five times in the dataset. Questions and answers are truncated to maximal lengths of 23 and 6, respectively. Since BERT representations are not convenient to adapt to the generation scenario, we use GloVe as the text representation for this experiment's methods. The video-question encoders are kept the same as in multi-choice QA. For answer decoders, we investigated several architectures; we found that GRU with soft attention over the questions performs well (see Appendix part 2 for details) and we use it for all models adapted from multi-choice QA. For better comparison, we also reproduce UATT \cite{xue2017unifying} which was proposed for generation-based open-ended QA by designing an order-preserved co-attention module.

\setlength{\belowcaptionskip}{-0.675cm}
\setlength{\tabcolsep}{13pt}
\begin{table}[t]
\begin{center}
\scalebox{0.7}{
\begin{tabular}{lccccc}
\toprule
Methods & $WUPS_C$ & $WUPS_T$ & $WUPS_D$ & $WUPS$ \cr
\midrule
\midrule
Popular & 9.73 & 8.95 & 28.39 & 13.40\cr
BlindQA & 12.14 & 14.85 & 40.41 & 18.88\cr
\midrule
STVQA \cite{jang2019VideoQA} & 12.52 & 14.57 & \underline{45.64} & 20.08\cr
HME \cite{fan2019heterogeneous} & 12.83 & 14.76 & 45.13 & 20.18\cr
HCRN \cite{le2020hierarchical} & 12.53 & \underline{15.37} & 45.29 & 20.25\cr
UATT \cite{xue2017unifying} & \underline{13.62} & {\bf16.23} & 43.41 & \underline{20.65}\cr
HGA \cite{jiang2020reasoning} & {\bf 14.76} & 14.90 & {\bf46.60} & {\bf21.48}\cr
\bottomrule
\end{tabular}
}
\caption{Results of open-ended QA on validation set.}
\label{tab:gqaresults-val}
\end{center}
\end{table}

\setlength{\belowcaptionskip}{-0.675cm}
\setlength{\tabcolsep}{13pt}
\begin{table}[t]
\begin{center}
\scalebox{0.7}{
\begin{tabular}{lccccc}
\toprule
Methods & $WUPS_C$ & $WUPS_T$ & $WUPS_D$ & $WUPS$ \cr
\midrule
\midrule
Popular & 12.19 & 10.79 & 31.94 & 16.12\cr
BlindQA & 14.87 & 18.35 & 45.78 & 22.66\cr
\midrule
STVQA \cite{jang2019VideoQA} & 15.24 & 18.03 & 47.11 & 23.04\cr
HCRN \cite{le2020hierarchical} & 16.05 & 17.68 & 49.78 & 23.92\cr
HME \cite{fan2019heterogeneous} & 15.78 & \underline{18.40} & \underline{50.03} & 24.06\cr
UATT \cite{xue2017unifying} & \underline{16.73} & {\bf18.68} & 48.42 & 24.25\cr
HGA \cite{jiang2020reasoning} & {\bf 17.98} & 17.95 & {\bf50.84} & {\bf25.18}\cr

\bottomrule
\end{tabular}
}
\caption{Results of open-ended QA on test set. We provide two reference answers for half of the test questions, and report the highest WUPS score between them.}
\label{tab:gqaresults-test}
\end{center}
\end{table}

\setlength{\belowcaptionskip}{-0.2cm}
\begin{figure*}[t]
    \centering
    \includegraphics[width=1.0\textwidth]{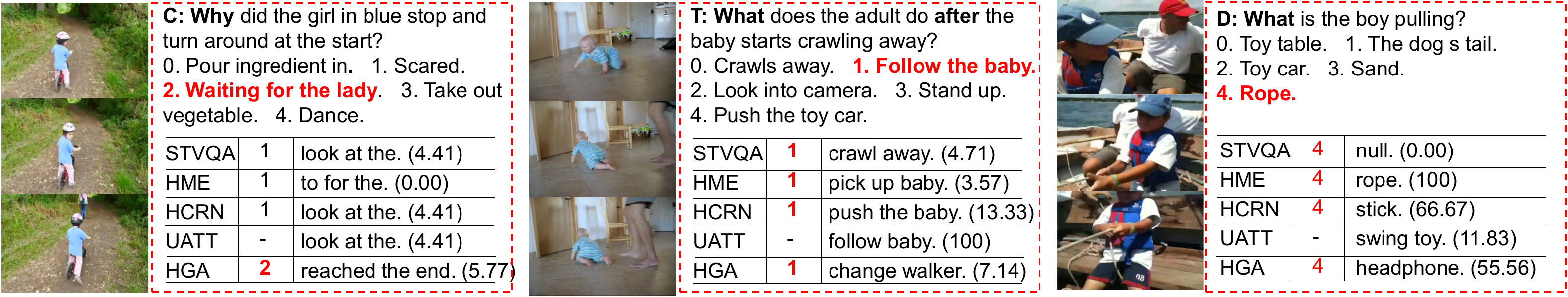}
    \caption{Visualization of answer prediction results. For multi-choice QA, the correct answers and predictions are highlighted in red. For open-ended QA, the WUPS score of each prediction is appended. 'null' means the methods fail to generate any effective words. (C: Causal. T: Temporal. D: Descriptive.)}
    \label{fig:vismcgqa}
\end{figure*}

\begin{figure}[t]
    \centering
    \includegraphics[width=0.48\textwidth]{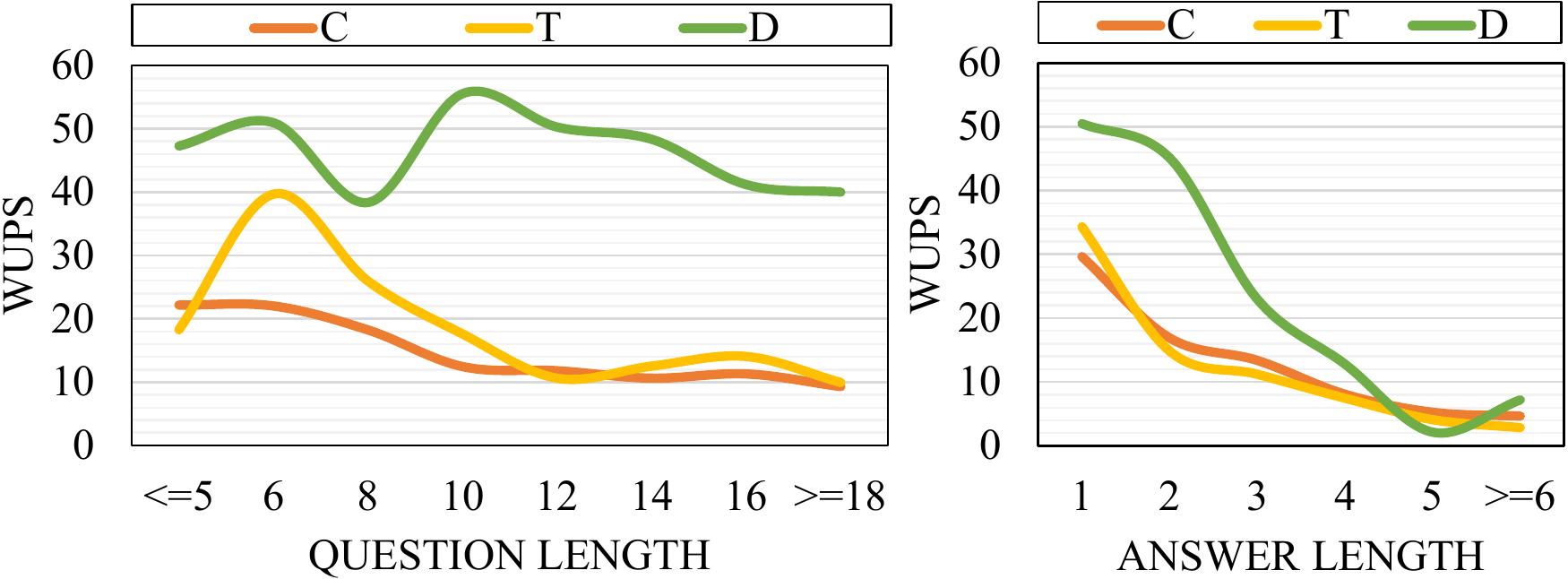}
    \caption{Result distribution on questions and answers.}
    \label{fig:qns-ans-length}
\end{figure}

As shown in Table \ref{tab:gqaresults-val} and Table \ref{tab:gqaresults-test}, although the methods can effectively boost the results over the BlindQA baseline, the overall improvements are trivial (less than 3\%) mainly due to the poor performance on causal and temporal questions. To delve into the reason, we first visualize some results in Figure \ref{fig:vismcgqa} (find more in Appendix part 3), from which we can see that the models struggle in automatically answering the questions, especially those which challenge causal and temporal action reasoning. We further detail the results of HGA \cite{jiang2020reasoning} (as a representative) on questions and answers of different lengths. As shown in Figure \ref{fig:qns-ans-length} (left), the performance on causal and temporal questions drops as the question lengths increase. However, for descriptive questions, the results are relatively stable and less impacted. Also, they are consistently better than causal and temporal questions. Regarding the answers in Figure \ref{fig:qns-ans-length} (right), the performances on all types of questions degrade on longer answers. By jointly considering the distributions of questions and answers in the dataset (refer to Figure \ref{fig:data-statistic}), we can draw that the models are essentially weak in causal and temporal reasoning and not strong enough for language understanding and generation.

\setlength{\belowcaptionskip}{-0.5cm}

\section{Discussion and Conclusion}
We conclude the following points and raise them as open challenges for the rest of the community. First, SOTA methods perform well on descriptive questions.  However, they are still weak in causal and temporal action reasoning -- the gap remains approximately 10\% and 30\% for multi-choice and open-ended QA respectively. Nonetheless, our empirical results suggest that graph models are superior for causal and temporal relation reasoning (refer to HGA \cite{jiang2020reasoning}) and are a promising direction to explore.
Regarding visual feature representations, motion feature are important but naively concatenating appearance and motion features usually results in sub-optimal results (refer to EVQA \cite{antol2015vqa}, PSAC+ \cite{li2019beyond} and STVQA \cite{li2016tgif}).  As such, we encourage investigating more effective ways of modelling and merging the two types of features. In terms of language representation, pre-trained BERT representations \cite{lu2019vilbert} are seriously biased to TextQA and generalize worse than that of GloVe \cite{pennington2014glove}. However, fine-tuned BERT shows absolute superiority in answering causal and temporal questions (refer to Tables \ref{tab:bs_results}, \ref{tab:mulresults}), and thus we recommend BERT as the text representation of choice for NExT-QA. Finally, we also find that it is better to contextualize (e.g., using RNN) than to directly operate on the pre-computed video/text features. 

Second, the methods that are effective on multi-choice QA struggle in automatically answering open-ended questions (see Tables \ref{tab:mulresults}, \ref{tab:sota_results} \vs Tables \ref{tab:gqaresults-val}, \ref{tab:gqaresults-test}; qualitative analysis in Figure \ref{fig:vismcgqa}). This prompts our fundamental concern that these methods do not truly understand the causal and temporal structures over actions. Instead, they are likely better at learning the differences between the provided correct and incorrect answers, which arguably, challenges more on grounding rather than inferring the answers in the videos \cite{xiao2020visual}. As such, we hope NExT-QA will underpin the next generation of VQA research not only in multi-choice QA, but also in open-ended QA.

Finally, open-ended QA is challenged not only by the reasoning component but also by language generation.
Our analysis shows that current VQA models are still weak in understanding complex questions and generating longer answers. Given the advancements made in vision-language representation learning \cite{sun2019videobert,Zhu_2020_CVPR}, future works are likely better served by using pre-trained architectures. Nevertheless, they need to be carefully balanced to incorporate and condition on the visual evidence. We believe this is also an exciting research area where NExT-QA can contribute towards advancements. Additionally, it could be beneficial to incorporate explicit relations, as NExT-QA's videos are sourced from VidOR \cite{shang2019annotating} where relation annotations already exist and provide a rich source of information to be leveraged.

\section*{Acknowledgement}
We greatly thank the reviewers for their positive remarks and some valuable suggestions. This research is supported by the National Research Foundation, Singapore under its International Research Centres in Singapore Funding Initiative, and under its NRF Fellowship for AI (NRF-NRFFAI1-2019-0001). Any opinions, findings and conclusions or recommendations expressed in this material are those of the author(s) and do not reflect the views of National Research Foundation, Singapore.

{\small
\bibliographystyle{ieee_fullname}
\bibliography{egbib}
}

\appendix
\section{NExT-QA Dataset}
\subsection{Data Statistics}
\begin{figure}[t]
\begin{center}
    \includegraphics[width=0.48\textwidth]{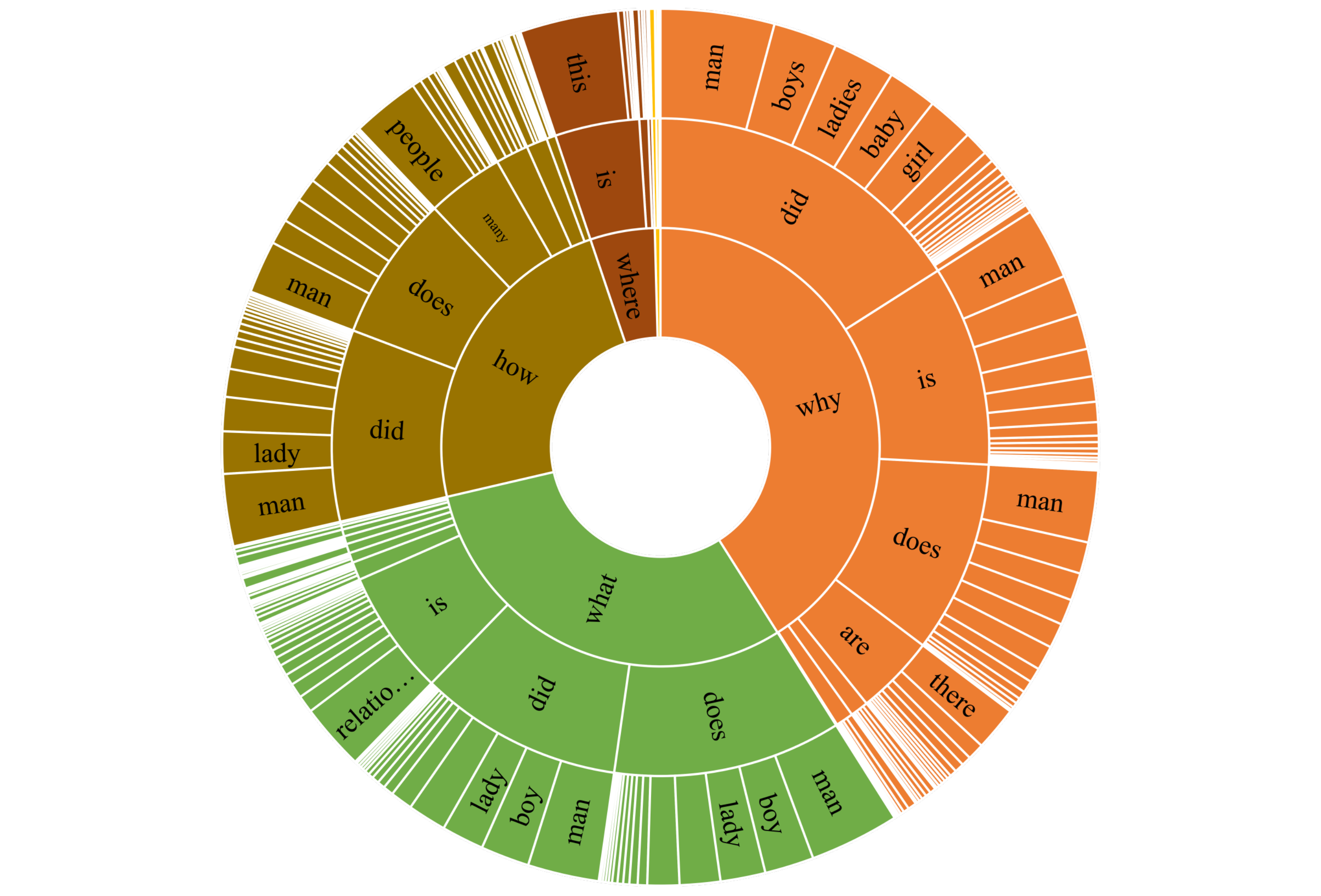}
    \caption{Distribution of NExT-QA questions by first three words. (The word ‘the’ in each question are ignored.)}
    \label{fig:qns-dis-word}
\end{center}
\end{figure}

\begin{figure*}[t]
    \centering
    \includegraphics[width=1.0\textwidth]{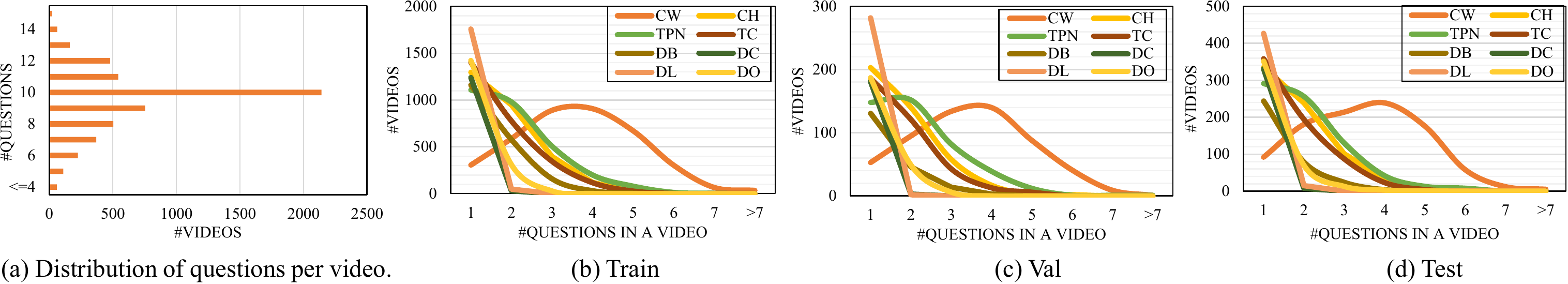}
    \caption{Distribution of the questions w.r.t videos. (a) the number of questions in most videos ranges from 4 to 14, and the vast majority of videos have 10 questions. In (b), (c) and (d), the distributions of questions are quite the same among the train/val/test data splits. For most of the videos, there are 2 to 6 causal questions that ask `why' (CW); 1 to 3 causal questions that ask `how' (CH); and 1 to 3 questions that ask temporal actions (either the previous/next (TPN) or the current (TC)). Aside from the causal and temporal questions, there are 1 or 2 descriptive questions asking either about binary-choice (DB), number-counting (DC), location (DL) or open-form (DO) in most of the videos.}
    \label{fig:qns-dis}
\end{figure*}

\begin{figure*}[t]
    \centering
    \includegraphics[width=0.8\textwidth]{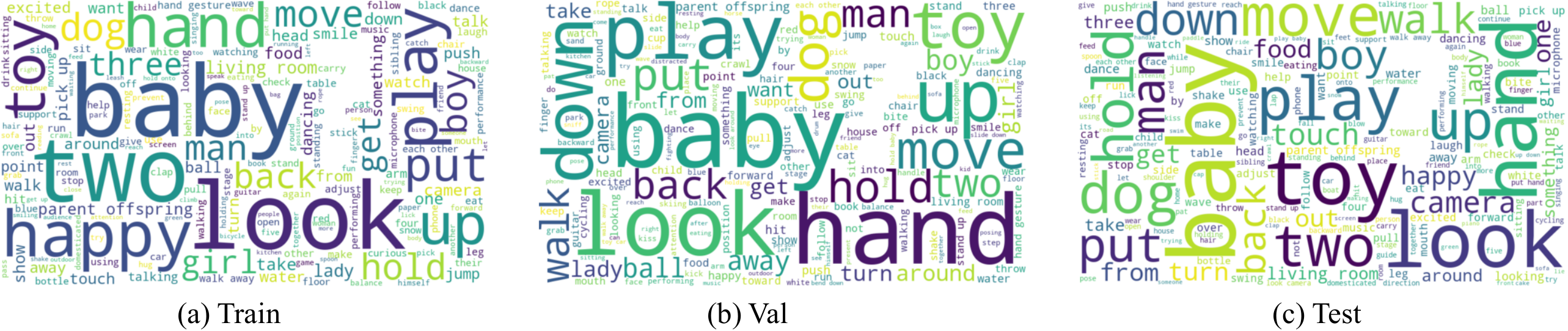}
    \caption{Word clouds for frequent words in answers (`yes', `no' and stop words are ignored.). The distributions vary little among train, val and test sets. This makes it possible to learn necessary information from training data for answering questions in val and test sets. Besides, the figures also show that there are various verbs in the answers in addition to nouns.}
    \label{fig:wordcloud-ans}
\end{figure*}

As shown in Figure \ref{fig:qns-dis-word}, questions in NExT-QA mostly ask \emph{'why did/does ...'}, and \emph{'how/what did/does ...'}. This reveals that NExT-QA advances existing VideoQA datasets that pay attention to scene recognition (\emph{what/who/where/which  is/are ...}) towards the explanation of temporal actions. The rich causal and temporal questions make NExT-QA a unique QA dataset for video understanding. Other details of the dataset are shown in Figure \ref{fig:qns-dis} and Figure \ref{fig:wordcloud-ans}.

\subsection{Dataset Comparison}
\begin{figure*}[t]
    \centering
    \includegraphics[width=1.0\textwidth]{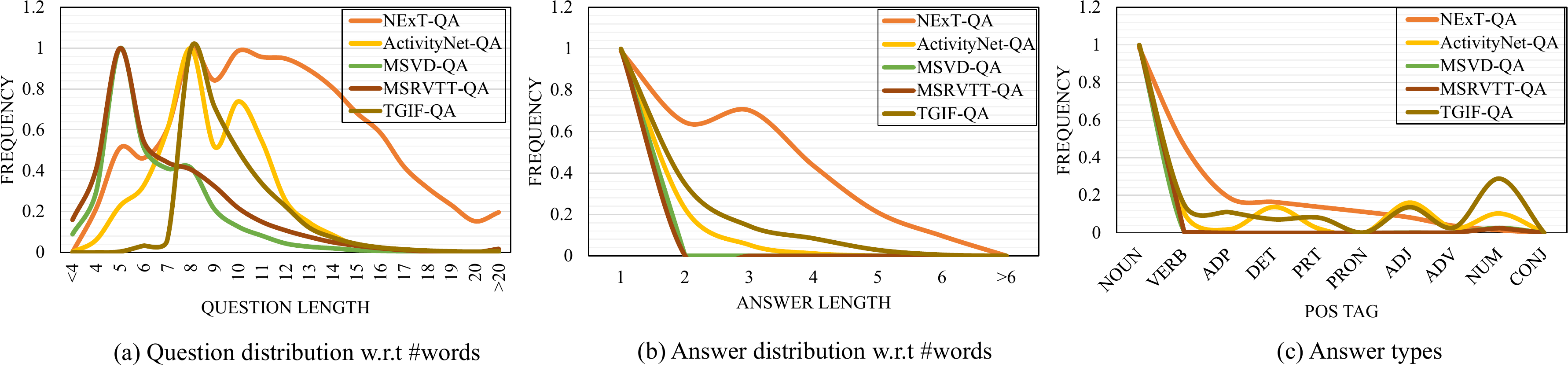}
    \caption{Detailed statistics of NExT-QA and popular VideoQA datasets.}
    \label{fig:comp-data}
\end{figure*}

In Figure \ref{fig:comp-data}, we compare NExT-QA with several related video QA datasets in terms of the distributions of QAs. From Figure \ref{fig:comp-data}(a) we can see that most of the questions in NExT-QA are relatively long, with an average of 12 words per question. 
Questions in MSVD-QA \cite{xu2017video} and MSRVTT-QA \cite{xu2017video} on the other hand are the shortest among the compared datasets (mostly 5 words). TGIF-QA \cite{jang2017tgif} and ActivityNet-QA \cite{yu2019activitynet} have about 8 words in most of their questions. Similarly, Figure \ref{fig:comp-data}(b) shows that the answers in NExT-QA are longer, with an average of 3 words, whereas TGIF-QA and ActivityNet-QA are dominated by one-word answers. In addition, we also find that all the answers in MSVD-QA and MSVTT-QA are with only one word. The relatively longer questions and answers in NExT-QA enable much more interesting QA contents, \ie, from recognition to explanation of video contents. In Figure \ref{fig:comp-data}(c), we show the frequency of the answer words in terms of their part-of-speech (POS) tags, from which we can see that the answers in NExT-QA are much richer in verbs because it focuses on the causal and temporal action reasoning. Though ActivityNet-QA and TGIF-QA explore temporal actions as well, they emphasize action recognition and object/repetition count. As a result, their answers are dominated by nouns and numbers. 

\begin{figure*}[t]
    \centering
    \includegraphics[width=0.9\textwidth]{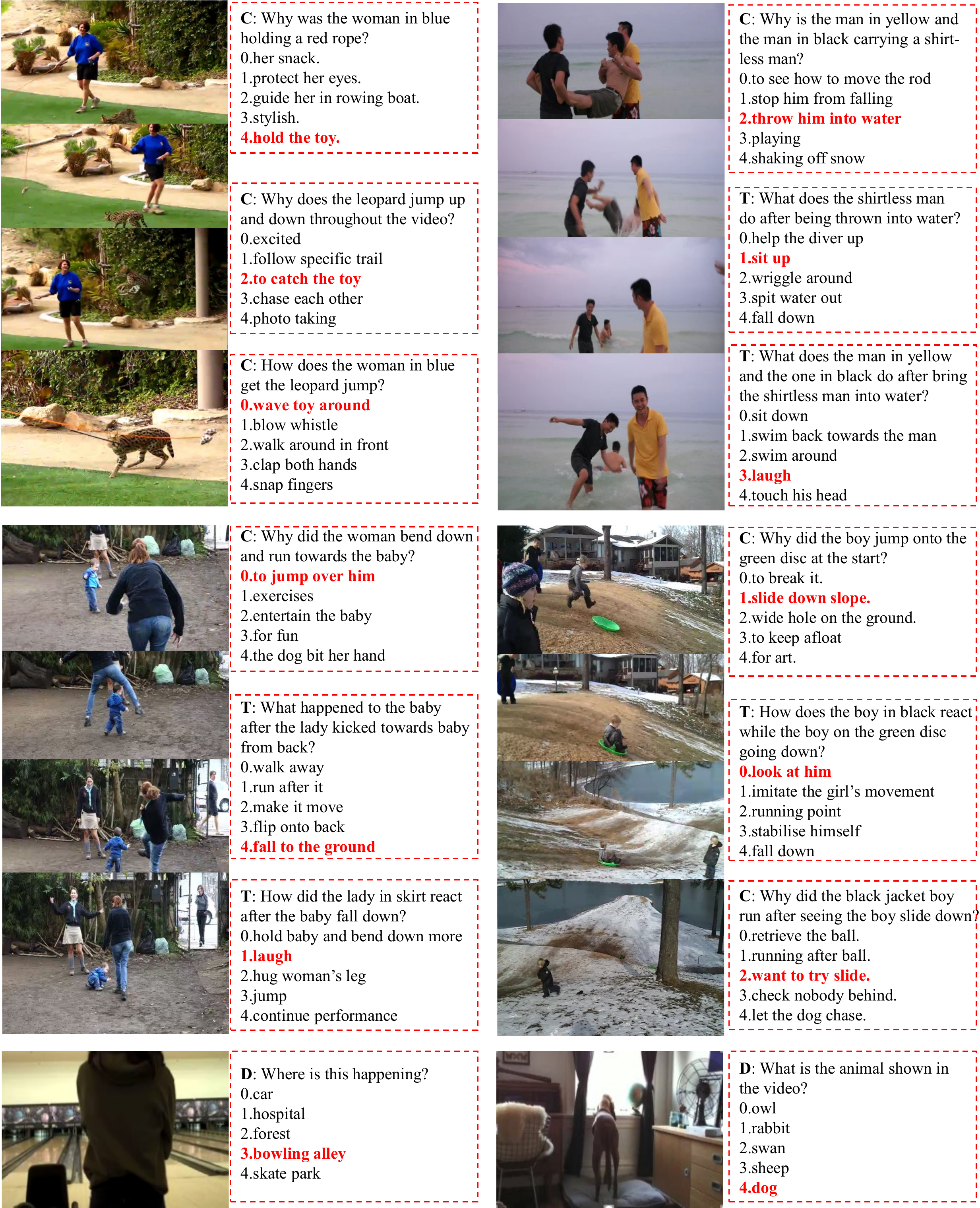}
    \caption{Examples of multi-choice QA. Each question has 5 options in which the correct answer is highlighted.}
    \label{fig:multi-exp}
\end{figure*}

The above statistical comparisons demonstrate that our NExT-QA dataset opens new challenges and opportunities for deeper understanding of video contents that goes beyond description. To better understand the dataset, we show some examples of the annotated question-answer pairs in Figure \ref{fig:vis-gqa} (open-ended QA) and Figure \ref{fig:multi-exp} (multi-choice QA), from which we can also confirm that the answers to the questions can be visually inferred from the video content.

\setlength{\belowcaptionskip}{-0.0cm}


\section{Analysis for open-ended QA}

\subsection{Evaluation}
WUPS score is introduced in \cite{malinowski2014multi} to evaluate the generated answers. It is regarded as a soft version of accuracy that factors in synonyms and semantics. Specifically, given a predicted answer $P = \{p_1, p_2, ..., p_i, ...\}$ for a question whose reference answer (ground truth) is $R = \{r_1, r_2, ..., r_i, ...\}$, in which $p_i$ and $r_i$ are the $i$th tokens of the predicted and reference answers respectively, the WUPS score computes the similarity between two token sets as follows:
\begin{equation}
\label{equ:wups}
\begin{aligned}
    WUPS(P, R) = \min\{\prod_{p\in P}\max \limits_{r\in R} WUP(p, r), \\
    \prod_{r\in R} \max \limits_{p\in P} WUP(r, p)\} \times 100,
\end{aligned}
\end{equation}
where $WUP(p, r)$ calculates the Wu-Parlmer similarity \cite{guadarrama2013youtube2text,wu1994verbs} of two words based on their depth in the taxonomy \cite{fellbaum2012wordnet,miller1995wordnet}: WUP(\emph{p}, \emph{r}) = 2*depth(\emph{lcs}) / (depth(\emph{p})  + depth(\emph{r})), in which \emph{lcs} is the least common ancestor of the words \emph{p} and \emph{r}. If two words are semantically closer, they would be in same/nearer depths in the hierarchy and share more common ancestors, and thus
get a higher WUP score.

\subsection{Answer Decoders}
\begin{figure*}[t]
    \centering
    \includegraphics[width=1.0\textwidth]{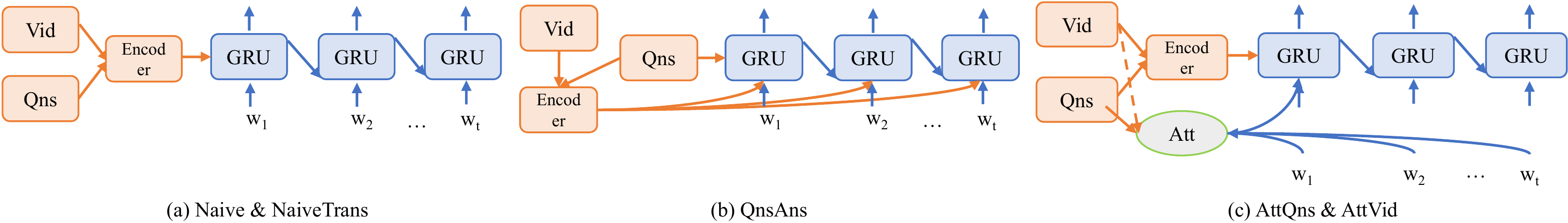}
    \caption{Architectures for answer decoders. (a) \emph{Naive} \cite{venugopalan2015sequence}, where the hidden state of the answer decoder is initialized with the output of the video-question (VQ) encoder. \emph{NaiveTrans} is a variant of \emph{Naive} by using transformation operation before RNNs. (b) \emph{QnsAns}, in which the hidden state of the decoder is initialized with the last hidden state of the question encoder, and the output of the VQ encoder is concatenated with the input of the decoder at each time step. (c) An attention variant of the naive implementation, in which the attention can either be added to the question (\emph{AttQns}) or the video (\emph{AttVid}) \cite{yao2015} side.}
    \label{fig:decoder}
\end{figure*}

For answer decoders, we investigated several architectures as shown in Figure \ref{fig:decoder}. The results in Table \ref{tab:decoder-results} are based on HGA \cite{jiang2020reasoning} on validation set. From the results, we can see that the \emph{naive} implementation performs the worst among other approaches. \emph{naiveTrans} achieves the best result on descriptive question, but it still struggles on causal and temporal questions. \emph{QnsAns} shows superior performance on temporal questions but is weak in answering causal questions featured in NExT-QA, and thus the overall WUPS score is still low. In contrast, \emph{AttVid} and \emph{AttQns} achieve better performances on causal questions and thus the better overall results. We attribute such strength of the attention-base decoders to the fact that they are better at determining which parts of the question or video should be attended for the answer. Since \emph{AttQns} achieved the best overall result, we choose it as the default answer decoders for all other methods adapted from multi-choice QA.

\setlength{\tabcolsep}{6.5pt}
\begin{table}[t]
\begin{center}
\scalebox{0.9}{
\begin{tabular}{lcccc}
\toprule
Methods & $WUPS_C$ & $WUPS_T$ & $WUPS_D$ & $WUPS$ \cr
\midrule
\midrule
Naive & 12.95 & 15.04 & 45.65 & 20.44\cr
NaiveTrans & 12.74 & 15.15 & {\bf47.58} & 20.77\cr
QnsAns & 12.50 & {\bf16.09} & \underline{46.82} & 20.77 \cr
AttVid & \underline{13.63} & 15.47 & 45.45 & \underline{20.85}\cr
AttQns & {\bf14.76} & 14.90 & 46.60 & {\bf21.48}\cr
\bottomrule
\end{tabular}
}
\caption{Results of different answer decoders.}
\label{tab:decoder-results}
\end{center}
\end{table}
\section{Results Analysis and Discussion}
\begin{figure*}[t]
    \centering
    \includegraphics[width=0.9\textwidth]{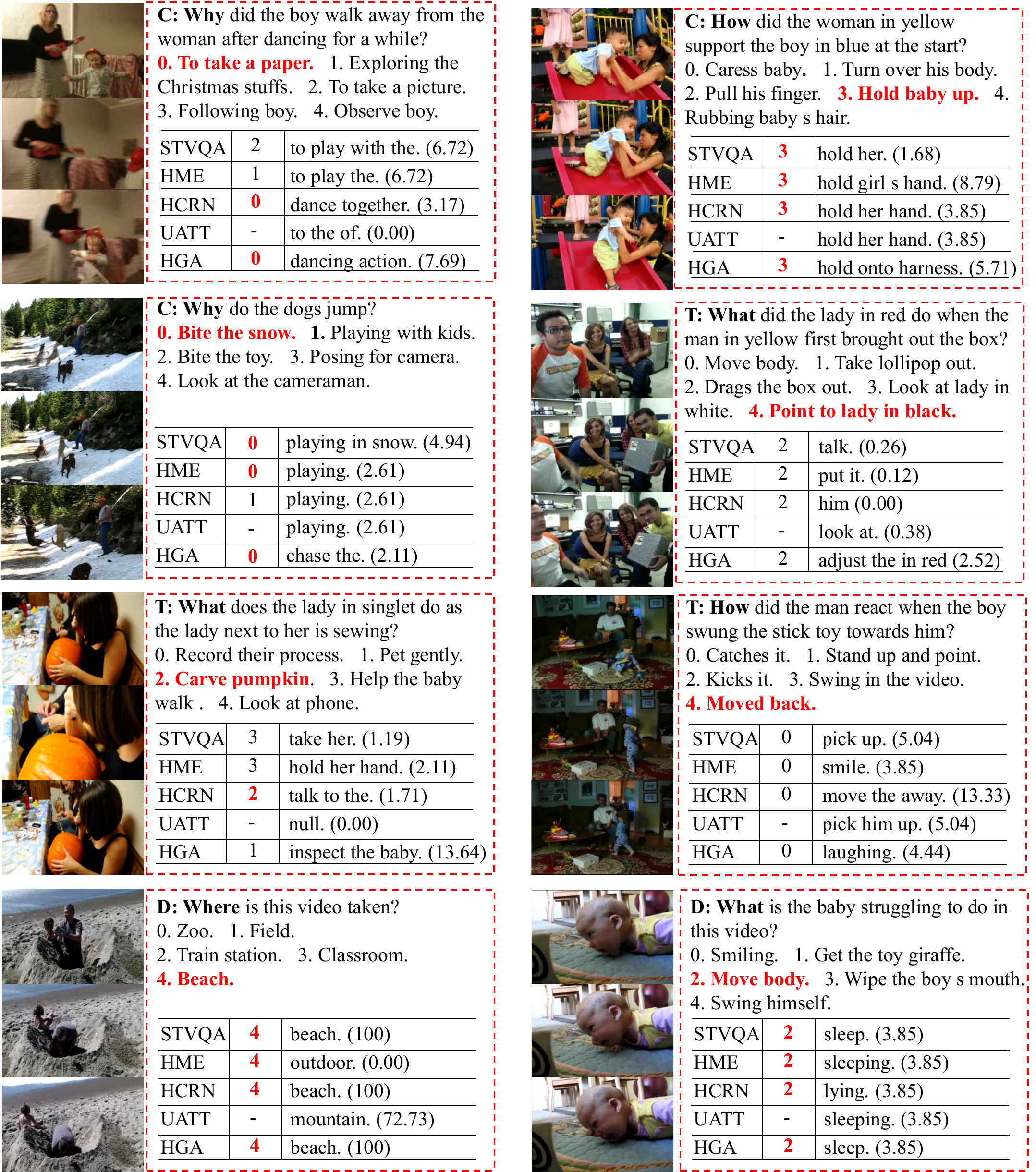}
    \caption{Visualization of answer predictions for both multi-choice QA and open-ended QA. For multi-choice QA, the correct answers and predictions are highlighted in red. For open-ended QA, the WUPS score of each prediction is appended. 'null' means the methods fail to generate any effective words. (C: Causal. T: Temporal. D: Descriptive.)}
    \label{fig:vis-gqa}
\end{figure*}

In Figure \ref{fig:vis-gqa}, we qualitatively analyze the models' performances on both multi-choice QA and open-ended QA tasks. According to the results, we make several main observations: 1) Answering causal and temporal questions requires much deeper understanding of both questions and videos that goes beyond a shallow description (refer to examples 1 to 6 \vs the last two), and the current models are still weak in this area. 2) When adapting models that are effective on multi-choice QA to open-ended QA, we find that they usually fail to correctly answer the questions, especially for causal and temporal questions (refer to examples 2, 3, 7 and 8). This suggest that the models either do not truly understand the video/questions, or struggle in generate the answers; both encumbering them from real-word application. 3) The models can correctly answer the questions to a certain extent in open-ended QA (refer to examples 2, 3 and 7), even if their WUPS scores are low with respect to the reference answers. 4) The predictions on some samples are semantically reasonable in answering the questions but are not relevant to the video contents (refer to examples 5, 6). This demonstrates that they can understand the questions, but are struggling in videos comprehension or language generation.

Overall, our NExT-QA dataset opens new challenges for deeper video understanding in that it benchmarks causal and temporal action reasoning, and is rich in object interactions in real-daily activities. Our extensive experiments show that existing models are weak in this area, which encourages future works for improvement.

\end{document}